\documentclass[conference]{IEEEtran}
\IEEEoverridecommandlockouts
\usepackage{cite}
\usepackage{amsmath,amssymb,amsfonts}
\usepackage{algorithmic}
\usepackage{graphicx}
\usepackage{textcomp}
\usepackage{xcolor}
\usepackage{multirow}
\usepackage{booktabs}
\usepackage{hyperref}
\hypersetup{
    colorlinks=true,
    linkcolor=blue,
    filecolor=magenta,      
    urlcolor=cyan,
    pdftitle={Overleaf Example},
    pdfpagemode=FullScreen,
    }
\def\BibTeX{{\rm B\kern-.05em{\sc i\kern-.025em b}\kern-.08em
    T\kern-.1667em\lower.7ex\hbox{E}\kern-.125emX}}
\begin{document}
\title{RHRSegNet: Relighting High-Resolution Night-Time Semantic Segmentation}

\author{\IEEEauthorblockN{ Sarah Elmahdy, Rodaina Hebishy, Ali Hamdi}
\IEEEauthorblockA{\textit{MSA University} \\
 Giza, Egypt \\
\{sarah.kamaleldeen, rodina.mohamed1,ahamdi\}@msa.edu.eg } }
 
\maketitle

\begin{abstract}
Night-time semantic segmentation is a crucial task in computer vision, focusing on accurately classifying and segmenting objects in low-light conditions. Unlike daytime techniques, which often perform worse in nighttime scenes, it is essential for autonomous driving due to insufficient lighting, low illumination, dynamic lighting, shadow effects, and reduced contrast. We propose RHRSegNet, implementing a relighting model over a High-Resolution Network for semantic segmentation. RHRSegNet implements residual convolutional feature learning to handle complex lighting conditions. Our model then feeds the lightened scene feature maps into a high-resolution network for scene segmentation. The network consists of a convolutional producing feature maps with varying resolutions, achieving different levels of resolution through down-sampling and up-sampling. Large nighttime datasets are used for training and evaluation, such as NightCity, City-Scape, and Dark-Zurich datasets. Our proposed model increases the HRnet segmentation performance by $5\%$ in low-light or nighttime images. The code is available at the following URL: \url{https://github.com/SarahELMAHDY03/RHRSegNet.git}.
\end{abstract}

\section{Introduction}
Semantic segmentation, a fundamental task in computer vision, involves the challenging process of assigning semantic labels to each pixel in an image. This significant effort enhances scene understanding and aids in recognizing objects within the visual context. Despite notable advances in semantic segmentation for daytime conditions, addressing the complexities of low-light conditions presents an additional set of challenges. With the growth of deep learning and machine learning, many techniques for semantic segmentation have recently been proposed. These methods perform very well; however, their application to nighttime imaging has not been extensively studied. Reduced visibility and increased noise in low-light conditions often cause existing approaches to fail in maintaining accuracy. This work aims to close this gap by introducing a novel strategy for semantic segmentation during nighttime. Utilizing deep learning techniques and algorithms adapted for low-light conditions, our goal is to enhance the precision and reliability of segmentation. An increasing number of researchers have started to segment more complex images under various types of degradation, such as those captured at night or in hazy conditions. Semantic segmentation of nighttime images is the main focus of this research, as it has numerous significant applications in autonomous driving. In this paper, we propose RHRSegNet, a model for nighttime semantic segmentation, using the Night-city, Cityscape, and Dark Zurich datasets. Previous work employed an intermediate twilight domain for gradually adapting semantic models trained in daylight scenarios to evening conditions \cite{dai2018dark}. \cite{sakaridis2019guided,sakaridis2020map} later expanded this into a guided curriculum adaptation framework, leveraging the cross-time-of-day relationship of scene photographs using both stylized synthetic images and unlabeled real images. However, these progressive adaptation methods typically require training multiple semantic segmentation models, such as the three models in \cite{sakaridis2019guided} for three distinct domains, which is highly inefficient. Moreover, their performance fails in some specific circumstances. Most existing semantic segmentation approaches are not designed for nighttime. While considerable progress has been made in semantic segmentation for daytime scenes, tackling low-light conditions presents unique challenges. This is particularly relevant in applications such as autonomous driving, where accurate scene interpretation at night is crucial for safe navigation. The primary contributions of our work are as follows: we propose RHRSegNet, which incorporates a relighting model followed by a segmentation model.
The main contributions of our work are summarized as follows:
\begin{itemize}
    \item We present RHRSegNet, a novel semantic segmentation method designed to address lighting challenges in nighttime environments, offering precise semantic information distinction in low-light conditions with minimal additional parameters.
    \item We improve nighttime semantic segmentation by transferring knowledge between the Cityscape dataset and Dark Zurich dataset, enhancing object segmentation in low-light situations by effectively transferring knowledge between domains.
    \item We experimented with RHRSegNet without domain adaptation on NightCity, achieving high performance.
\end{itemize}
The rest of the paper is structured as follows. Section $2$ introduces the related work. Section $3$ presents the proposed research methodology. The datasets used and experimental results are explained in Sections $4$ and $5$, respectively.
\section{Related Work}
\subsection{Nighttime Semantic Segmentation}
Semantic segmentation at night has seen advancements through various methods. \cite{dai2018dark} introduced the concept of using an intermediate twilight domain to gradually transform semantic models trained in daylight scenarios to evening situations. This approach was expanded upon by \cite{sakaridis2019guided,sakaridis2020map}, who developed a guided curriculum adaptation framework leveraging the cross-time-of-day relationship in scene photographs, employing both stylized synthetic images and unlabeled actual images. However, such progressive adaptation methods often necessitate training multiple semantic segmentation models for different domains, as seen in \cite{sakaridis2019guided}. This approach can be resource-intensive and wasteful. Subsequent research \cite{romera2019bridging,sun2019see,nag2019s} has explored the use of additional image transfer models, such as CycleGAN \cite{zhu2017unpaired}, for day-to-night or night-to-day image translation prior to training semantic segmentation models. The effectiveness of these techniques heavily relies on the performance of the pre-trained image transfer model. Recognizing the resilience of thermal radiation to changes in light, Vertens et al. \cite{vertens2020heatnet} proposed the incorporation of thermal infrared images alongside RGB images for nighttime semantic segmentation. Additionally, domain adaptation techniques, as demonstrated in \cite{di2020rainy}, have been employed for semantic segmentation of night scenes. NightLab \cite{deng2022nightlab} categorizes items as simple or complex. A Hardness Detection Module identifies difficult categories and sends them to a segmentation module. \cite{xie2023boosting} suggest employing image frequency distributions for nighttime scene analysis. However, these methods do not estimate the impact of lighting on semantics.
\subsection{Semantic Segmentation}
Semantic segmentation aims to categorize each pixel individually. Recently, methods based on Fully Convolutional Networks (FCN) \cite{long2015fully} paired with encoder-decoder architectures have dominated this field. Various techniques \cite{chen2014semantic,chen2017deeplab,yu2015multi} utilize dilated convolutions to expand the receptive field, while PSPNet \cite{zhao2017pyramid} employs a pyramid pooling module (PPM) to capture multi-scale contexts. Integrating these advancements, the DeepLab series \cite{chen2014semantic,chen2017deeplab,chen2017rethinking,chen2018encoder} introduces atrous spatial pyramid pooling (ASPP) to incorporate contextual information. More recently, self-attention mechanisms have been extensively used to capture long-range dependencies in semantic segmentation. Additionally, several transformer-based networks, including Vision Transformer and Swin Transformer, have been used to provide stronger backbones. Moreover, ConvNeXt \cite{liu2022convnet} has emerged as a competitive alternative to transformers. However, the majority of these methods are primarily focused on daytime scenes.

\section{Research Methodology}
\subsection{Relighting Model}
\begin{figure*}[h]
    \centering
    \includegraphics[width=.7\textwidth]{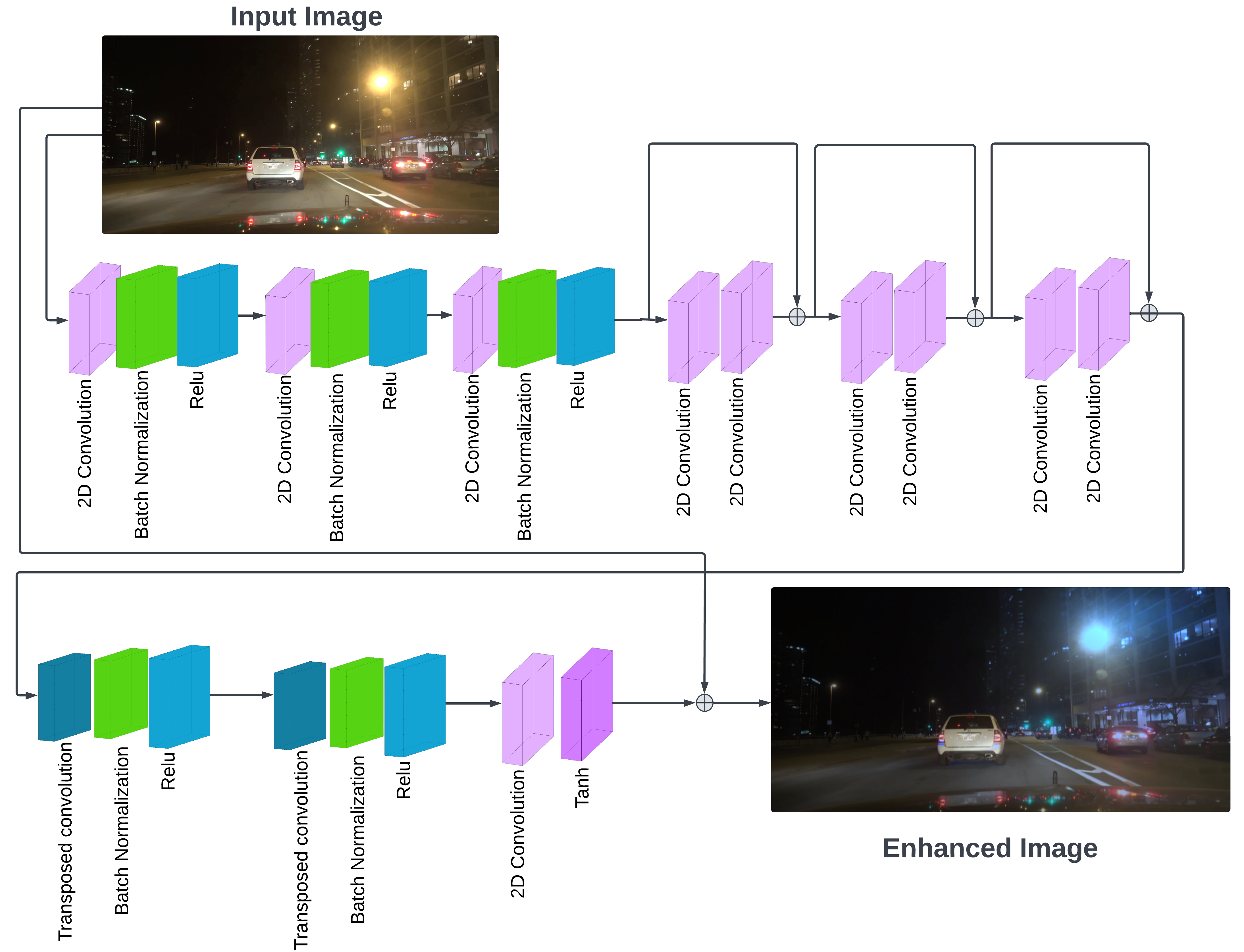}
    \caption{Relighting Model}
    \label{fig:1}
\end{figure*}
The relighting model simulates lighting changes using a shared-weight architecture, allowing for a robust and unified approach. The relighting model takes an input image, passes it through four convolutional layers, three residual blocks, and two transposed convolutional layers. Each layer is followed by a batch normalization layer. The relighted image is created by combining the output of the last layer with the input image. The resulting relighted image is then passed to the HRNet.
\begin{equation}
I_{\text{e}} = \text{BN1}(\text{TransConv2D}(\text{BN2}(\text{Conv2D}(I))) + I)
\end{equation}
In this simplified equation:
\(I_{\text{e}}\) represents the enhanced image. \\
\(\text{Conv1}\) and \(\text{Conv2}\) are convolutional layers. \\
\(\text{BN1}\) and \(\text{BN2}\) are batch normalization layers. \\
\(+\) denotes the skip connection, where \(I\) is added to the output of the first convolutional layer before being processed by the second convolutional layer and batch normalization, as shown in Figure \ref{fig:1}.
\subsection{Learning Parameter Sharing}
This research utilizes shared weights in the relighting network to optimize learning parameter utilization, improve computational and memory efficiency, generate transferable insights, and ensure consistent feature extraction. Let \( W \) represent the shared weights within the relighting network. Let \( f(x) \) represent the function of the relighting network, where \( x \) is the input image. Then, the output of the relighting network can be represented as:
\begin{equation}
    \text{Output} = f(x; W)
\end{equation}
Here, \( f(x; W) \) denotes the function \( f \) parameterized by the shared weights \( W \), which generates the relighted output given the input image \( x \). The equation explains how shared weights are used in a relighting network to adapt to different lighting conditions and input domains, enhancing computation, memory utilization, and generalization capabilities. This approach improves knowledge transfer across datasets and generalization capabilities, outperforming traditional methods in computational efficiency and result quality. It also optimizes model training and inference processes, allowing the network to quickly adjust to new lighting conditions.
\subsection{Semantic Segmentation Network}
\begin{figure*}[h]
    \centering
    \includegraphics[width=.7\textwidth]{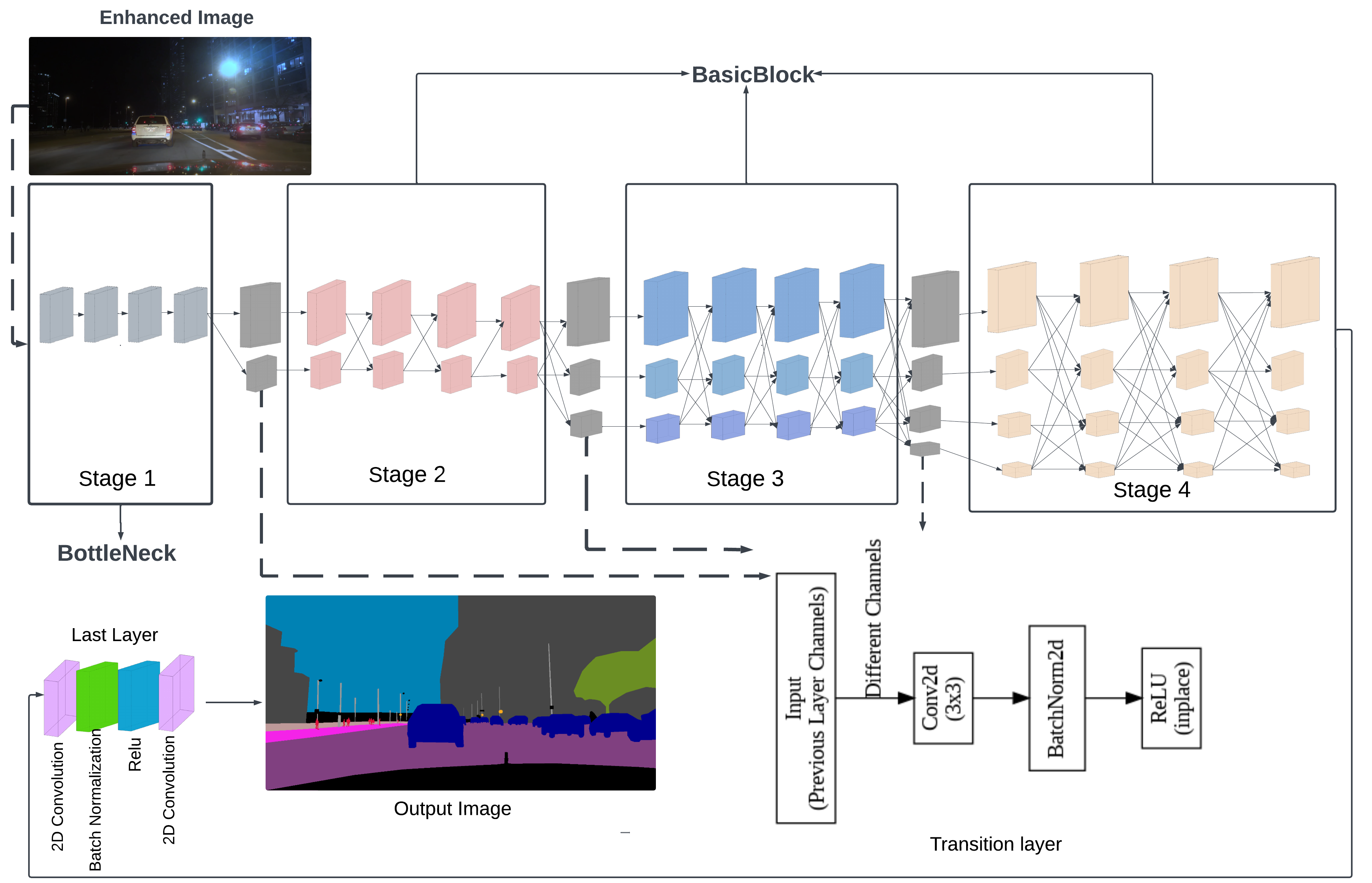}
    \caption{Our HRNet Architecture}
    \label{fig: Figure 2}
\end{figure*}
High-resolution representations are crucial for high-resolution problems like semantic segmentation, human pose estimation, and object detection. The High-Resolution Network (HRNet) maintains these representations by connecting high-to-low resolution convolution streams in parallel and repeatedly exchanging information across resolutions. The HRNet architecture is divided into many phases, each of which aims to incrementally improve and aggregate multi-scale characteristics for accurate semantic segmentation. Our model consists of a series of fundamental blocks, starting with an input layer, followed by convolutional layers, Batch Normalization, ReLU activation functions, residual connections, and an output layer. The model uses a hierarchical processing pipeline, which includes downsampling, convolutional operations, Batch Normalization, ReLU activation functions, and an output layer for model predictions.
This model takes the relighted image and processes the input features at the original resolution, setting the model at the first stage (STAGE1). Then, by adding more branches and blocks, the subsequent stages (STAGE2, STAGE3, STAGE4) increase the network's capacity and allow it to capture features at various resolutions. The transition layer is used to bridge feature maps of different resolutions between the adjacent stages of the network. Each phase consists of high-resolution modules that combine feature maps from many branches using techniques such as interpolation and summation to provide a comprehensive representation.
\begin{equation}
\text{Output} = \text{Concatenation}(F1, F2, F3, F4)
\end{equation}
where:
\begin{itemize}
    \item \( F1 \) represents the feature map extracted from stage 1
    \item \( F2 \) represents the feature map extracted from stage 2
    \item \( F3 \) represents the feature map extracted from stage 3
    \item \( F4 \) represents the feature map extracted from stage 4
\end{itemize}
The output from each stage from the 4 stages are processed through two $1\times1$ convolutions, followed by batch normalization and ReLU activation, and finally producing the output with 19 channels, as shown in Figure \ref{fig: Figure 2}.
\subsection{Data Augmentation}
Image transformations are crucial for data preparation in deep learning. Input images are converted into tensors and normalized to ensure consistent values, stability, and effective model training. Our data augmentation process includes cropping, resizing, and flipping to enhance dataset diversity and improve model generalization. The dataset's segmentation masks are transformed into tensor representations.The transformations are applied to the incoming data without issue, preparing it for further analysis and model training. This process ensures the model's success across various contexts.
\section{Datasets}
\subsection{Night-city}
Night City \cite{tan2021night,deng2022nightlab} is the largest available nighttime semantic segmentation dataset. It comprises $2,998$ images at a resolution of $1024 \times 512$ for training and $1,299$ images for validation. Each image comes with pixel-level annotations. However, there are noticeable annotation errors, such as missing labels and incorrectly labeled areas, which reduce the dataset's effectiveness. Additionally, \cite{wei2023disentangle} introduced NightCity-fine, an improved dataset derived from Night City for nighttime semantic segmentation. In this dataset, unreasonable annotations have been systematically corrected in both the training and validation sets.
\subsection{Dark-Zurich}
The Dark-Zurich dataset, a popular benchmark in computer vision and robotics research, is a collection of high-resolution images, primarily used for object detection and semantic segmentation tasks. The dataset features high-quality images with high spatial resolution, enabling detailed examination of urban environments. It includes a diverse range of urban scenes such as roads, crosswalks, buildings, green spaces, and other infrastructure. Additionally, many images are manually annotated with pixel-level labels, providing crucial ground truth annotations for semantic segmentation tasks. The dataset is well-suited for testing, comprising $8,377$ images for training and $100$ images for validation, evenly divided into $50$ night and $50$ day images. The test set includes 302 images, split into subsets for nighttime and daytime scenes, respectively.
\subsection{Cityscape}
The Cityscapes dataset is a comprehensive resource for urban scene understanding, featuring $5,000$ high-resolution frames captured across diverse street scenes. It includes pixel-level annotations across 19 semantic categories, such as cars, pedestrians, roads, traffic signs, and buildings. The dataset's landscape orientation ensures consistency and smooth integration into computer vision pipelines. With a resolution of $2,048 \times 1,024$ pixels, the images provide sufficient detail for accurate semantic segmentation and scene understanding tasks. Divided into three subsets, the dataset offers sample data for model training, validation, and testing, with $2,975$ images allocated for training, $500$ images for validation, and $1,525$ images for testing.
\subsection{Experimental Design}
We initiated our training by utilizing domain adaptation techniques on the Cityscapes and Dark Zurich datasets. Domain adaptation methods were employed to enhance the model's performance on Dark Zurich images. Specifically, we utilized techniques such as adversarial training to align the feature distributions between the two domains, using the SGD optimizer. By adapting the model trained on Cityscapes to better generalize to Dark Zurich, our goal was to minimize the domain shift and improve the model's robustness in low-light conditions. The results are shown in Table~\ref{tab:CS_DZ}. Additionally, we experimented with RHRSegnet on the NightCity-Fine dataset for training and obtained extremely good performance metrics without using domain adaptation approaches. Despite the lack of domain adaptation, the model showed proficiency in correctly classifying objects in night settings. This achievement demonstrates the adaptability and flexibility of RHRSegnet to changing environmental factors, especially low light levels, without requiring further domain-specific changes. The remarkable outcomes suggest that RHRSegnet has the ability to generalize effectively over a wide range of datasets, highlighting its potential for practical use in tasks involving the analysis and segmentation of night scenes. Results evaluated on the Cityscapes dataset are shown in Table~\ref{tab:CS}.
\subsection{Results}
The benchmark results presented in Table~\ref{tab:CS_DZ} offer a comparison of different segmentation models based on their mean intersection over union (mIOU) performance in the context of Cityscapes $\rightarrow$ Dark Zurich adaptation, evaluated on the Dark Zurich validation set. The first model, AdaptSegNet, employs the Deeplab architecture and achieves a modest mIOU of $20.2$. The next two models, GCMA and MGCDA, utilize the RefineNet architecture, yielding mIOUs of $26.7$ and $26.1$, respectively, showcasing a notable improvement over AdaptSegNet. Notably, our model, RHRSegNet, stands out with the highest mIOU of $27.19$, leveraging the HRNet architecture. This performance underscores the effectiveness of our model in handling domain adaptation challenges, particularly in challenging low-light conditions as represented in the Dark Zurich dataset. The results clearly highlight the superior performance of RHRSegNet, suggesting that our chosen architecture and model optimizations effectively enhance segmentation accuracy, outperforming the other competitive models in this challenging task.
\begin{table}[h]
    \centering
    \begin{tabular}{lcc}
        \toprule
        Method & Architecture & mIOU \\ 
        \midrule
        AdaptSegNet \cite{tsai2018learning} & Deeplab & 20.2 \\ 
        GCMA \cite{sakaridis2019guided} & RefineNet & 26.7 \\ 
        MGCDA \cite{sakaridis2020map} & RefineNet & 26.1 \\ 
        RHRSegNet (Ours) & HRNet & 27.19 \\ 
        \bottomrule
    \end{tabular}
    \vspace{0.2cm}
    \caption{\textnormal{Comparison by mIOU metric using Cityscapes $\rightarrow$ Dark Zurich adaptation evaluated on Dark Zurich validation between our model RHRSegNet and other models}}
    \label{tab:CS_DZ}
\end{table}
The benchmark results in Table~\ref{tab:CS} showcase a comparison of various segmentation models based on their mean intersection over union (mIoU) performance using the Cityscapes dataset for evaluation. RefineNet, an advanced segmentation network, achieves a respectable mIoU of $28.5$. DeepLab-v2 and PSPNet, both well-known models in the domain, slightly outperform RefineNet, each with an mIoU of $28.8$. However, our model, RHRSegNet, demonstrates a substantial improvement, achieving an impressive mIoU of $37.53$. This result highlights the effectiveness of RHRSegNet in segmenting complex urban scenes, significantly outperforming other models. The remarkable performance of RHRSegNet underscores its robust architecture and superior feature extraction capabilities, making it highly effective for semantic segmentation tasks, especially within urban environments as represented in the Cityscapes dataset.

\begin{table}[h]
    \centering
    \begin{tabular}{lc}
        \toprule
        Method & mIoU \\ 
        \midrule
        RefineNet \cite{li2019bidirectional} & 28.5 \\ 
        DeepLab-v2 \cite{chen2017deeplab} & 28.8 \\ 
        PSPNet \cite{zhao2017pyramid} & 28.8 \\ 
        RHRSegNet & 37.53 \\ 
        \bottomrule
    \end{tabular}
    \vspace{0.2cm}
    \caption{\textnormal {Comparison by mIoU metric using Cityscapes for evaluation}}
    \label{tab:CS}
\end{table}

\begin{table*}[h]
    \centering
    \resizebox{\textwidth}{!}{%
    \begin{tabular}{lllllllllllllllllllll}
        \toprule
        Method & \rotatebox{90}{Road} & \rotatebox{90}{Sidewalk} & \rotatebox{90}{Building} & \rotatebox{90}{Wall} & \rotatebox{90}{Fence} & \rotatebox{90}{Pole} & \rotatebox{90}{Traffic Light} & \rotatebox{90}{Traffic Sign} & \rotatebox{90}{Vegetation} & \rotatebox{90}{Terrain} & \rotatebox{90}{Sky} & \rotatebox{90}{Person} & \rotatebox{90}{Rider} & \rotatebox{90}{Car} & \rotatebox{90}{Truck} & \rotatebox{90}{Bus} & \rotatebox{90}{Train} & \rotatebox{90}{Motorcycle} & \rotatebox{90}{Bicycle} & mIoU \\ \midrule
        RefineNet \cite{li2019bidirectional} & 68.8 & 23.2 & 46.8 & 20.8 & 12.6 & 29.8 & 30.4 & 26.9 & 43.1 & 14.3 & 0.3 & 36.9 & 49.7 & 63.6 & 6.8 & 0.2 & 24.0 & 33.6 & 9.3 & 28.5 \\
        DeepLab-v2 \cite{chen2017deeplab} & 79.0 & 21.8 & 53.0 & 13.3 & 11.2 & 22.5 & 20.2 & 22.1 & 43.5 & 10.4 & 18.0 & 37.4 & 33.8 & 64.1 & 6.4 & 0.0 & 52.3 & 30.4 & 7.4 & 28.8 \\
        PSPNet \cite{zhao2017pyramid} & 78.2 & 19.0 & 51.2 & 15.5 & 10.6 & 30.3 & 28.9 & 22.0 & 56.7 & 13.3 & 20.8 & 38.2 & 21.8 & 52.1 & 1.6 & 0.0 & 53.2 & 23.2 & 10.7 & 28.8 \\ \midrule
        AdaptSegNet \cite{tsai2018learning} & 86.1 & 44.2 & 55.1 & 22.2 & 4.8 & 21.1 & 5.6 & 16.7 & 37.2 & 8.4 & 1.2 & 35.9 & 26.7 & 68.2 & 45.1 & 0.0 & 50.1 & 33.9 & 15.6 & 30.4 \\
        ADVENT \cite{vu2019advent} & 85.8 & 37.9 & 55.5 & 27.7 & 14.5 & 23.1 & 14.0 & 21.1 & 32.1 & 8.7 & 2.0 & 39.9 & 16.6 & 64.0 & 13.8 & 0.0 & 58.8 & 28.5 & 20.7 & 29.7 \\
        BDL \cite{li2019bidirectional} & 85.3 & 41.1 & 61.9 & 32.7 & 17.4 & 20.6 & 11.4 & 21.3 & 29.4 & 8.9 & 1.1 & 37.4 & 22.1 & 63.2 & 28.2 & 0.0 & 47.7 & 39.4 & 15.7 & 30.8 \\
        DMAda \cite{lee2018diverse} & 75.5 & 29.1 & 48.6 & 21.3 & 14.3 & 34.3 & 36.8 & 29.9 & 49.4 & 13.8 & 0.4 & 43.3 & 50.2 & 69.4 & 18.4 & 0.0 & 27.6 & 34.9 & 11.9 & 32.1 \\ \midrule
        RHRSegNet-Cityscapes (Ours) & 92.59 & 58.56 & 78.65 & 5.83 & 10.11 & 30.48 & 16.33 & 32.45 & 78.97 & 30.35 & 80.25 & 52.98 & 14.57 & 80.77 & 0.46 & 0.02 & 0.02 & 0.11 & 49.52 & 37.53 \\
        RHRSegNet-NightCity-Fine (Ours) & 87.25 & 37.63 & 77.84 & 23.79 & 32.72 & 23.58 & 9.87 & 28.06 & 49.62 & 14.63 & 82.82 & 29.76 & 0.01 & 70.84 & 20.65 & 24.55 & 0.0 & 0.0 & 20.01 & 33.35 \\
        \bottomrule
    \end{tabular}%
    }
    \vspace{0.1cm}
    \caption{\textnormal{Comparison by 19 Classes and mIoU} }
    \label{tab:CCS}
\end{table*}
The benchmark results presented in Table~\ref{tab:CCS} provide a comprehensive comparison of segmentation models across various categories based on their mean Intersection over Union (mIoU) performance using the Cityscapes and Dark Zurich datasets. Notably, the models evaluated include RefineNet, DeepLab-v2, PSPNet, AdaptSegNet, ADVENT, BDL, DMAda, and our proposed RHRSegNet with two variations: RHRSegNet-Cityscapes and RHRSegNet-NightCity-Fine. In terms of overall performance, RHRSegNet-Cityscapes clearly outperforms the other models with a remarkable mIoU of $37.53$, showcasing exceptional segmentation accuracy, particularly in key categories like Road, Building, Sky, Car, and Bus. This reflects the robust architecture and superior feature extraction capabilities of RHRSegNet. The second variation, RHRSegNet-NightCity-Fine, also performs competitively with an mIoU of $33.35$, highlighting its effectiveness in challenging low-light conditions.
Among the competing models, DMAda demonstrates strong performance with an mIoU of $32.1$, particularly excelling in categories like Rider and Car. Meanwhile, BDL achieves an mIoU of $30.8$, showing solid results across several categories. The other models, including RefineNet, DeepLab-v2, PSPNet, AdaptSegNet, and ADVENT, achieve mIoUs below $30$, indicating that they struggle more with the segmentation tasks, particularly in challenging urban and nighttime environments.
Overall, the results highlight the superior performance of RHRSegNet, particularly in complex and challenging conditions, reaffirming its robust design and superior segmentation capabilities across a wide range of urban scene categories.
\begin{figure*}[h]
    \centering
    \begin{minipage}{0.30\textwidth}
        \centering
        \includegraphics[width=\textwidth]{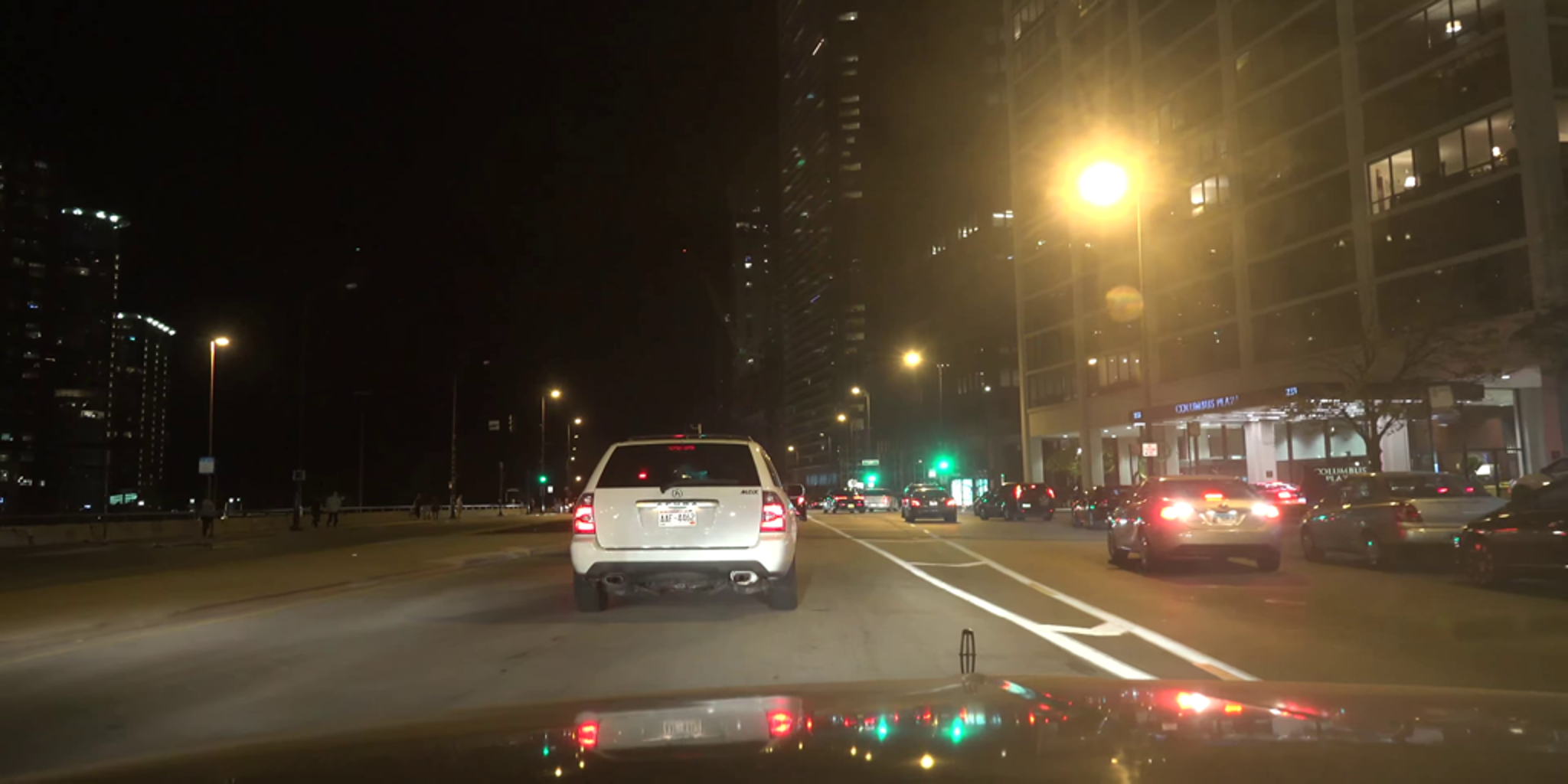}
        \caption{Original Image}
    \end{minipage}\hfill
    \begin{minipage}{0.30\textwidth}
        \centering
        \includegraphics[width=\textwidth]{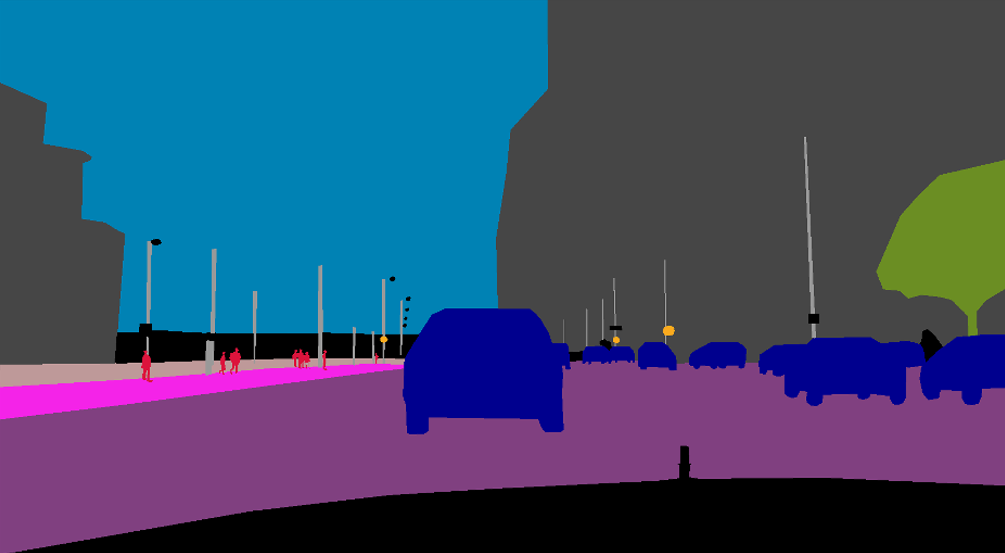}
        \caption{Ground Truth}
    \end{minipage}\hfill
    \begin{minipage}{0.30\textwidth}
        \centering
        \includegraphics[width=\textwidth]{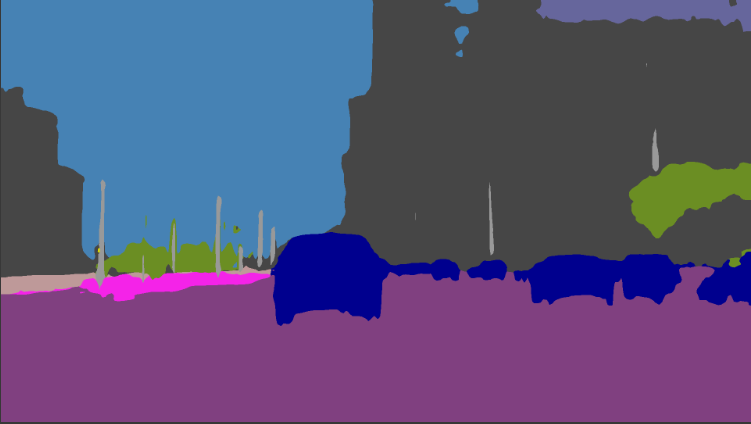}
        \caption{Our RHRSegNet}
    \end{minipage}
\end{figure*}
\section{Conclusion}
The paper presents a novel model, RHRSegNet, for nighttime semantic segmentation using deep learning techniques and low-light-adapted algorithms. The model evenly distributes light, simulates changes in lighting conditions to improve brightness and contrast, and separates different factors in an image, such as lighting conditions and object colors. It also distinguishes between different features in an image, thereby improving image quality and processing images with multiple variables. The model is tested using the Night-City, Cityscape, and Dark-Zurich datasets, demonstrating its potential in handling complex lighting conditions and recognizing semantics even in challenging scenarios. This approach sets a stronger standard for evaluating segmentation throughout the night, enhancing the quality of computer vision applications like autonomous driving.
\bibliographystyle{plain}
\bibliography{main}

\begin{thebibliography}{10}

\bibitem{chen2014semantic}
Liang-Chieh Chen, George Papandreou, Iasonas Kokkinos, Kevin Murphy, and Alan~L Yuille.
\newblock Semantic image segmentation with deep convolutional nets and fully connected crfs.
\newblock {\em arXiv preprint arXiv:1412.7062}, 2014.

\bibitem{chen2017deeplab}
Liang-Chieh Chen, George Papandreou, Iasonas Kokkinos, Kevin Murphy, and Alan~L Yuille.
\newblock Deeplab: Semantic image segmentation with deep convolutional nets, atrous convolution, and fully connected crfs.
\newblock {\em IEEE transactions on pattern analysis and machine intelligence}, 40(4):834--848, 2017.

\bibitem{chen2017rethinking}
Liang-Chieh Chen, George Papandreou, Florian Schroff, and Hartwig Adam.
\newblock Rethinking atrous convolution for semantic image segmentation.
\newblock {\em arXiv preprint arXiv:1706.05587}, 2017.

\bibitem{chen2018encoder}
Liang-Chieh Chen, Yukun Zhu, George Papandreou, Florian Schroff, and Hartwig Adam.
\newblock Encoder-decoder with atrous separable convolution for semantic image segmentation.
\newblock In {\em Proceedings of the European conference on computer vision (ECCV)}, pages 801--818, 2018.

\bibitem{dai2018dark}
Dengxin Dai and Luc Van~Gool.
\newblock Dark model adaptation: Semantic image segmentation from daytime to nighttime.
\newblock In {\em 2018 21st International Conference on Intelligent Transportation Systems (ITSC)}, pages 3819--3824. IEEE, 2018.

\bibitem{deng2022nightlab}
Xueqing Deng, Peng Wang, Xiaochen Lian, and Shawn Newsam.
\newblock Nightlab: A dual-level architecture with hardness detection for segmentation at night.
\newblock In {\em Proceedings of the IEEE/CVF Conference on Computer Vision and Pattern Recognition}, pages 16938--16948, 2022.

\bibitem{di2020rainy}
Shuai Di, Qi~Feng, Chun-Guang Li, Mei Zhang, Honggang Zhang, Semir Elezovikj, Chiu~C Tan, and Haibin Ling.
\newblock Rainy night scene understanding with near scene semantic adaptation.
\newblock {\em IEEE Transactions on Intelligent Transportation Systems}, 22(3):1594--1602, 2020.

\bibitem{lee2018diverse}
Hsin-Ying Lee, Hung-Yu Tseng, Jia-Bin Huang, Maneesh Singh, and Ming-Hsuan Yang.
\newblock Diverse image-to-image translation via disentangled representations.
\newblock In {\em Proceedings of the European conference on computer vision (ECCV)}, pages 35--51, 2018.

\bibitem{li2019bidirectional}
Yunsheng Li, Lu~Yuan, and Nuno Vasconcelos.
\newblock Bidirectional learning for domain adaptation of semantic segmentation.
\newblock In {\em Proceedings of the IEEE/CVF conference on computer vision and pattern recognition}, pages 6936--6945, 2019.

\bibitem{liu2022convnet}
Zhuang Liu, Hanzi Mao, Chao-Yuan Wu, Christoph Feichtenhofer, Trevor Darrell, and Saining Xie.
\newblock A convnet for the 2020s.
\newblock In {\em Proceedings of the IEEE/CVF conference on computer vision and pattern recognition}, pages 11976--11986, 2022.

\bibitem{long2015fully}
Jonathan Long, Evan Shelhamer, and Trevor Darrell.
\newblock Fully convolutional networks for semantic segmentation.
\newblock In {\em Proceedings of the IEEE conference on computer vision and pattern recognition}, pages 3431--3440, 2015.

\bibitem{nag2019s}
Sauradip Nag, Saptakatha Adak, and Sukhendu Das.
\newblock What’s there in the dark.
\newblock In {\em 2019 IEEE International Conference on Image Processing (ICIP)}, pages 2996--3000. IEEE, 2019.

\bibitem{romera2019bridging}
Eduardo Romera, Luis~M Bergasa, Kailun Yang, Jose~M Alvarez, and Rafael Barea.
\newblock Bridging the day and night domain gap for semantic segmentation.
\newblock In {\em 2019 IEEE Intelligent Vehicles Symposium (IV)}, pages 1312--1318. IEEE, 2019.

\bibitem{sakaridis2019guided}
Christos Sakaridis, Dengxin Dai, and Luc~Van Gool.
\newblock Guided curriculum model adaptation and uncertainty-aware evaluation for semantic nighttime image segmentation.
\newblock In {\em Proceedings of the IEEE/CVF International Conference on Computer Vision}, pages 7374--7383, 2019.

\bibitem{sakaridis2020map}
Christos Sakaridis, Dengxin Dai, and Luc Van~Gool.
\newblock Map-guided curriculum domain adaptation and uncertainty-aware evaluation for semantic nighttime image segmentation.
\newblock {\em IEEE Transactions on Pattern Analysis and Machine Intelligence}, 44(6):3139--3153, 2020.

\bibitem{sun2019see}
Lei Sun, Kaiwei Wang, Kailun Yang, and Kaite Xiang.
\newblock See clearer at night: towards robust nighttime semantic segmentation through day-night image conversion.
\newblock In {\em Artificial Intelligence and Machine Learning in Defense Applications}, volume 11169, pages 77--89. SPIE, 2019.

\bibitem{tan2021night}
Xin Tan, Ke~Xu, Ying Cao, Yiheng Zhang, Lizhuang Ma, and Rynson~WH Lau.
\newblock Night-time scene parsing with a large real dataset.
\newblock {\em IEEE Transactions on Image Processing}, 30:9085--9098, 2021.

\bibitem{tsai2018learning}
Yi-Hsuan Tsai, Wei-Chih Hung, Samuel Schulter, Kihyuk Sohn, Ming-Hsuan Yang, and Manmohan Chandraker.
\newblock Learning to adapt structured output space for semantic segmentation.
\newblock In {\em Proceedings of the IEEE conference on computer vision and pattern recognition}, pages 7472--7481, 2018.

\bibitem{vertens2020heatnet}
Johan Vertens, Jannik Z{\"u}rn, and Wolfram Burgard.
\newblock Heatnet: Bridging the day-night domain gap in semantic segmentation with thermal images.
\newblock In {\em 2020 IEEE/RSJ International Conference on Intelligent Robots and Systems (IROS)}, pages 8461--8468. IEEE, 2020.

\bibitem{vu2019advent}
Tuan-Hung Vu, Himalaya Jain, Maxime Bucher, Matthieu Cord, and Patrick P{\'e}rez.
\newblock Advent: Adversarial entropy minimization for domain adaptation in semantic segmentation.
\newblock In {\em Proceedings of the IEEE/CVF conference on computer vision and pattern recognition}, pages 2517--2526, 2019.

\bibitem{wei2023disentangle}
Zhixiang Wei, Lin Chen, Tao Tu, Pengyang Ling, Huaian Chen, and Yi~Jin.
\newblock Disentangle then parse: Night-time semantic segmentation with illumination disentanglement.
\newblock In {\em Proceedings of the IEEE/CVF International Conference on Computer Vision}, pages 21593--21603, 2023.

\bibitem{xie2023boosting}
Zhifeng Xie, Sen Wang, Ke~Xu, Zhizhong Zhang, Xin Tan, Yuan Xie, and Lizhuang Ma.
\newblock Boosting night-time scene parsing with learnable frequency.
\newblock {\em IEEE Transactions on Image Processing}, 2023.

\bibitem{yu2015multi}
Fisher Yu and Vladlen Koltun.
\newblock Multi-scale context aggregation by dilated convolutions.
\newblock {\em arXiv preprint arXiv:1511.07122}, 2015.

\bibitem{zhao2017pyramid}
Hengshuang Zhao, Jianping Shi, Xiaojuan Qi, Xiaogang Wang, and Jiaya Jia.
\newblock Pyramid scene parsing network.
\newblock In {\em Proceedings of the IEEE conference on computer vision and pattern recognition}, pages 2881--2890, 2017.

\bibitem{zhu2017unpaired}
Jun-Yan Zhu, Taesung Park, Phillip Isola, and Alexei~A Efros.
\newblock Unpaired image-to-image translation using cycle-consistent adversarial networks.
\newblock In {\em Proceedings of the IEEE international conference on computer vision}, pages 2223--2232, 2017.

\end{thebibliography}
\end{document}